# Analysis of Biomass Sustainability Indicators from a Machine Learning Perspective

**Syeda Nyma Ferdous[1], Xin Li[1], Kamalakanta Sahoo [2], Richard Bergman[2]**

[1]West Virginia University, Morgantown, WV

[2] Forest Products Laboratory, Forest Service, United States Department of Agriculture, Madison, WI USA

**\* Corresponding Author: sf0070@mix.wvu.edu**

## Abstract

Plant biomass estimation is critical due to the variability of different environmental factors and crop management practices associated with it. The assessment is largely impacted by the accurate prediction of different environmental sustainability indicators. A robust model to predict sustainability indicators is a must for the biomass community. This study proposes a robust model for biomass sustainability prediction by analyzing sustainability indicators using machine learning models. The prospect of ensemble learning was also investigated to analyze the regression problem. All experiments were carried out on a crop residue data from the Ohio state. Ten machine learning models, namely, linear regression, ridge regression, multilayer perceptron, k-nearest neighbors, support vector machine, decision tree, gradient boosting, random forest, stacking and voting, were analyzed to estimate three biomass sustainability indicators, namely soil erosion factor, soil conditioning index, and organic matter factor. The performance of the model was assessed using cross-correlation ($R^2$), root mean squared error and mean absolute error metrics. The results showed that Random Forest was the best performing model to assess sustainability indicators. The analyzed model can now serve as a guide for assessing sustainability indicators in real time.

## 1    Introduction

Sustainable plant biomass is one of the most prominent bioenergy resources [1][2][3]. Biomass production is highly dependent on the accurate estimation of different indicators of environmental sustainability indicators [4][5]. The most important and widely used sustainability indicators are the soil erosion factor (SEF), the soil conditioning index (SCI), and the organic matter factor (OMF) related to bioenergy applications. These factors are used to assess soil quality, which is a major consideration for siting of biomass plants. Machine learning models can be used to accurately quantify different sustainability indicators so that informed decisions can be made for biomass production [6].

Optimal biomass estimation is an active research area. Gleason and Lm [7] conducted a study on forest biomass estimation using machine learning approaches. To predict biomass gasification, machine learning methods have been explored in [8]. Zhou et al. [9] estimated wheat biomass using a machine learning model and Sahoo et al. [10] estimated cotton stalks in the Southeastern US using an artificial neural network (ANN) model. Sahoo et al. [11] performed a predictive model based on the geographic information system (GIS) to estimate sustainable crop residues for optimal plant location. Biogas plants. A detailed study by Morais et al. [12] on grassland estimation using machine learning methods suggested that the use of machine learning along with remote sensing data can improve the performance of biomass estimation.

Assessing biomass from grid-level data in the GIS platform is both time-consuming and memory-hungry. Machine learning models are a great alternative to designing the prediction system, as the time complexity and memory requirement are significantly reduced for these models with excellent accuracy. The installation of biomass plants is also affected by the estimation of sustainability indicators. Therefore, our proposed approach can aid investors and policy makers in the decision-making process for the development of biomass plants [13].

Ensemble Learning [14] [15]is a machine learning technique that is used to combine several base models into one for robust prediction performance. Ensemble learning involves different techniques such as bagging, boosting, voting, and stacking. Bagging fits many decision trees on different samples of the dataset and averages the prediction, while boosting arranges models sequentially where predictions made by prior models have corrected and a weighted average of the prediction is computed. Voting combines different machine learning models and returns the average of the predicted values. Stacking fits different base models on the same data and then uses a final model to best combine the prediction.

In this study, we analyzed the potential of different machine learning models for estimating different indicators of biomass sustainability. We have also explored the potential of ensemble learning to solve this regression problem. The model performance is verified using standard goodness-of-fit metrics (cross-correlation ($R^2$), root mean squared error (RMSE), mean absolute error (MAE)) [16]. The results suggest that the proposed method can be applied for the successful estimation of biomass.

## 2    Case Study

In this study, the applicability of different machine learning models was investigated for the primary sustainability indicators of biomass production. The analysis was performed on crop residue data collected from Ohio state. Different ensemble models have been developed to estimate sustainability indicators relevant to biomass production (e.g. SEF, SCI, and OMF) with the highest correlation coefficient ($R^2$) and lowest standard error value. Soil-related information (e.g., soil type, texture, slope length, etc.) was collected from the Geographic Gridded Soil Survey (gSSURGO) database. Furthermore, the RUSLE2 tool was used to generate synthetic SEF, SCI, and OMF values for the study area. The input variables to predict sustainability indicators include biomass yield (corn, soybean, wheat), soil characteristics (soil erodibility, soil texture, soil organic matter content), rainfall erosivity, slope (soil surface inclined relative to the horizontal), slope length, K-factor (microbial biomass potassium), crop rotations, and residue removal rate. The availability of sustainable crop residue depends on the studied input features. The dataset features were represented in Fig. 1.

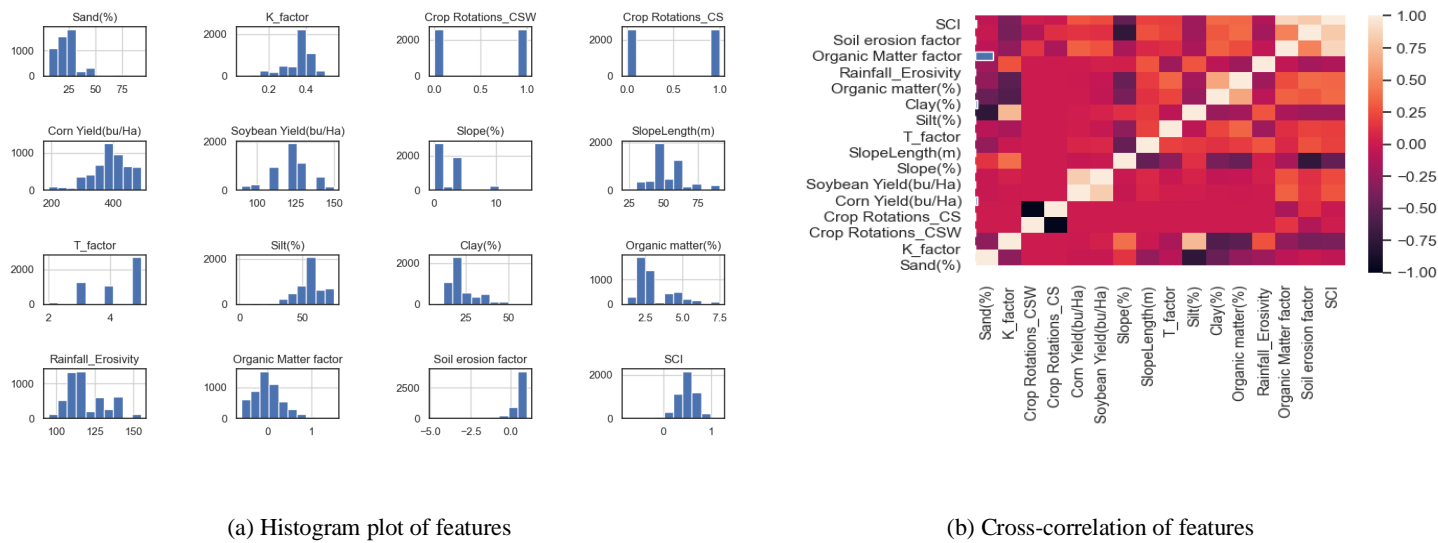

(a) Histogram plot of features        (b) Cross-correlation of features

Figure 1: Feature Visualization of the Ohio dataset. Here, histogram showing features on an interval scale while cross-correlation matrix showing different correlations between features.

## 3    Models

In this study, we have analyzed different ensemble learning approaches together with independent models to solve the regression problem.

### 3.1    Linear Regression

Linear regression (LR) [17][18] is a supervised machine learning algorithm that assumes a linear relationship between a dependent variable and one or more independent variables. Also, LR assumes that the independent variables would not be co-linear which constrain the model to generate linear solutions. The model tries to predict the unknown parameters estimated from the data linearly. The objective of this algorithm is to fit a line to the data points that minimizes the deviation of the data points from the regression line. LR uses mean squared error (MSE) as the loss function to minimize this deviation.

### 3.2    Ridge regression

Ridge regression [19][20]is a method for estimating the coefficients of multiple regression models in scenarios where independent variables are highly correlated. It was developed as a possible solution to the imprecision of least-square estimators when linear regression models have some multicollinear (highly correlated) independent variables. Ridge regression tries to regularize linear regression with $l2$ penalty. Ridge regression achieves a more precise estimate of ridge parameters (e.g., alpha that controls $l2$ regularization, maximum number of iterations, normalize) than the linear regression counterpart.

### 3.3    Multi-Layer Perceptron

Multi-layer perceptron (MLP), one of the widely studied models for real-world data, is used to model nonlinear relationships underlying practical data [21] [22]. These models are comprised of nodes interconnected to one another via three different layers, namely the input layer, hidden layer, and output layer. The inputs first passes through the input layer. Then, the hidden layer performs a series of transformations to compute the hidden representation of features. Finally, the output layer provides the analyzed outputs of the model. MLP computes the weighted sum of the inputs and includes a bias for each training sample. The weighted total is passed through an activation function which is responsible for producing the final output.

### 3.4   k-Nearest neighbors

The k nearest neighbors (kNN) [23] [24] can be used for classification and regression. In both cases, the input consists of the k closest training examples in the data set. In a classification setting, the output is a class membership. An object is classified by a majority vote of its k nearest neighbors. The value of k can be predefined or can be assessed by cross -validation. In the regression setting, the output is the property value for the object obtained by the average of the values of k nearest neighbors. It uses different distance functions such as Euclidean, Manhattan, Minkowski for distance measure between data points.

### 3.5   Support Vector Regression

The Support Vector Regression (SVR) [25] [26] is a regression algorithm that handles regression as a convex optimization problem. SVR tries to find the hyperplane by estimating the best fit line which captures maximum number of data points in a higher-dimensional feature space. The model tries to find the optimal hyperplane within a threshold value or error margin ($\mathcal{C}$) for improved performance. In SVR, threshold value ($\mathcal{C}$) is the distance between the hyperplane and boundary line. To map the data points in higher dimension, it uses the kernel trick. Kernel trick uses different kernel functions (Gaussian kernel, Radial Basis Function (RBF) kernel, etc.) for modeling optimal separation of nonlinear data points.

### 3.6 Decision Tree

Decision trees (DTs), as the name implies are tree-structured nonparametric supervised learning methods [27] [28]. These methods work by learning decision rules to solve both classification and regression problems. For regression, a decision tree tries to fit to the target variable considering each of the independent variables. The decision tree is split into branches for each independent variable. The objective is to minimize the sum of squared residuals. To select the best features for splitting, some popular metrics (e.g., information gain, Entropy, Gini impurity) have been devised.

### 3.7 Gradient Boosting

Gradient boost gives a prediction model in the form of a set of weak prediction models, which are typically decision trees [29][30]. The term gradient is associated with using gradient descent algorithm for minimizing the loss function. When a decision tree is a weak learner, the resulting algorithm is called gradient-boosted trees; it usually outperforms random forests.

### 3.8 Random Forest

Random Forest [31] is an ensemble-based supervised learning model in which a collection of uncorrelated decision tree models is trained in parallel using the bagging principle. Random Forest takes decisions considering all the combinations of the predicted results of individual decision tree outcomes, which helps to achieve good generalization performance in terms of solving regression problems. In random forest, the training samples are selected randomly while performing splitting which prevents overfitting problem. This algorithm performs well on large datasets and dataset with missing data points.

### 3.9 Voting

Voting [32] is a metamodel based on the ensemble that predicts by averaging the models. In voting, the performance of each model is taken into consideration and then model averaging is performed to compute the final prediction performance. As voting requires multiples models to perform final performance estimation, this technique is computationally expensive.

### 3.10 Stacking

Stacking or Stacked Generalization [33] is a group machine learning algorithm that uses a meta-learning method to learn how to best combine the predictions of multiple base machine learning algorithms. The benefit of stacking is that it can harness the capabilities of several well-performing models on either classification or regression tasks, which achieves predictions with better performance than any single model in the ensemble.

### 4. Model Training and Testing

Initially, all 16 variables shown in Fig. 1 were used as input features for the model. Later, backward elimination was used to remove insignificant features. The data set was divided into two subsets: 80% train and 20% test. The data set was normalized in such a way that each feature had a mean of 0 and a standard deviation of 1. The optimum parameter was chosen by observing the cross-validation error of the data points.

The performance of the multilayer perceptron (MLP) model depends on several configuration factors such as the number of layers, learning rate, transfer function, etc. Higher performance can be achieved with a complex MLP architecture (e.g., increasing the number of layers and the number of nodes in hidden layers). However, complex MLP architecture is not feasible for the problem domain, as it increases time complexity with greater usage of memory.

The performance of kNN varies with the selection of the number of neighbors which is shown in Fig. 2(a). The performance of SVM is highly dependent on the appropriate kernel selection. With the help of the kernel function, the non-linear problem was solved linearly. We experimented with three types of SVM kernel: sigmoid, RBF kernel, and polynomial (Fig. 2(b)). As shown in Fig. 2(c), RBF and polynomial kernel gave almost similar performance to predict SCI, SEF, and OMF. The reason can be the nonlinear nature of our data points. The RF performance for the differing number of trees is reported in Fig. 2 (c). Although the SCI prediction value was not significantly affected by the number of trees in the RF model, the prediction performance of SEF and OMF varied with the specified parameter.

Performance is estimated based on $R^2$, RMSE and MAE. $R^2$ is the goodness-of-measure for how a model fits the data. $R^2$ was computed using Eq.1. RMSE and MAE were computed using Eq. 2 and 3 respectively. Since in RMSE, errors are squared, giving relatively large weights to higher errors compared to MAE.

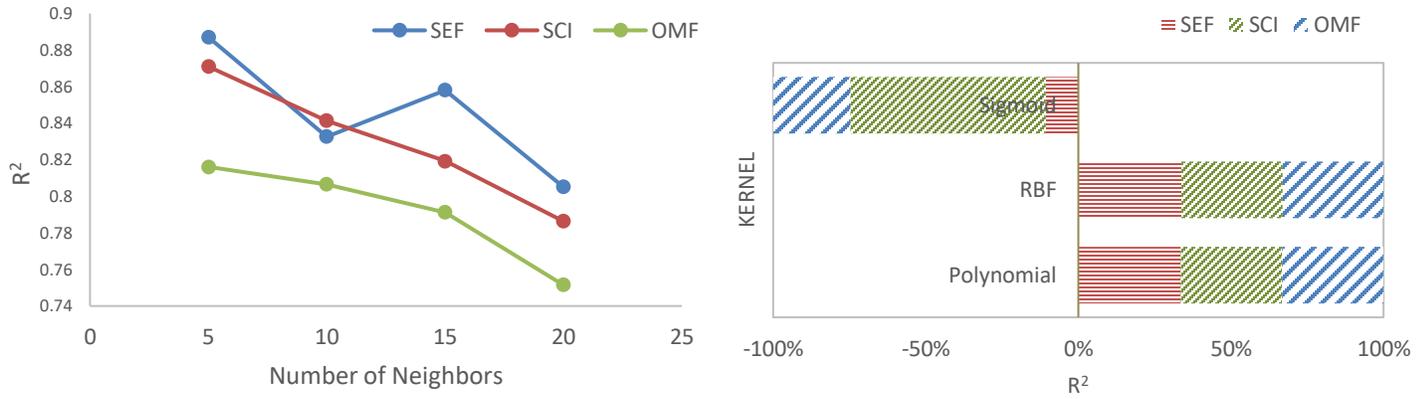

**(a)** k-nearest neighbors performance with different number of neighbors

**(b)** Support Vector Machine performance with different kernel

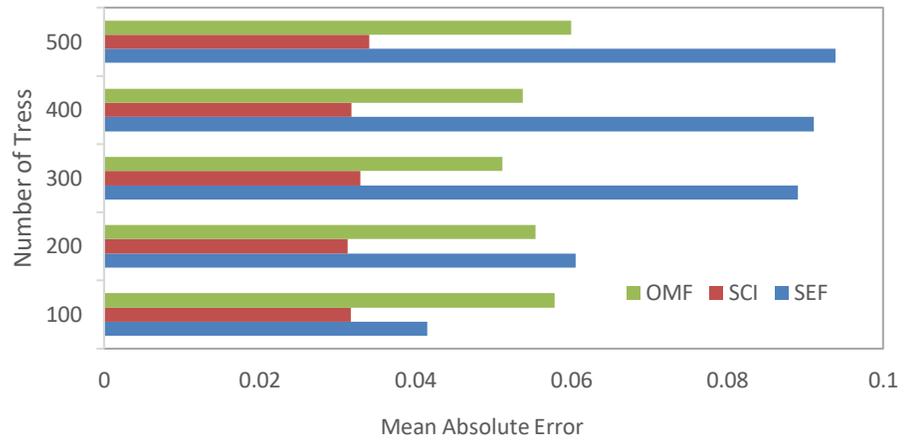

**(c)** Random Forest performance for different numbers of estimator trees

Figure 2: Different performance of the machine learning model for varying parameters. SEF: Soil Erosion Factor; SCI: Soil Conditioning Index; OMF: Organic Matter Factor

Higher $R^2$ along with reduced MAE and RMSE ensures a good-performing model.

$$R^2 = \frac{\sum_{i=1}^{n}(S_{predicted} - \overline{S_{predicted}})\,(S_{actual} - \overline{S_{actual}})}{\sqrt{\sum_{i=1}^{n}(S_{predicted} - \overline{S_{predicted}})^2}\,\sqrt{\sum_{i=1}^{n}(S_{actual} - \overline{S_{actual}})^2}} \tag{1}$$

$$RMSE = \sqrt{\frac{1}{n}\sum_{i=1}^{n}(S_{predicted} - S_{actual})^2} \tag{2}$$

$$MAE = \frac{1}{n}\sum_{i=1}^{n}||\,S_{predicted} - S_{actual}\,|| \tag{3}$$

where n is the number of samples, $S_{actual}$ and $S_{predicted}$ are ground truth and predicted data, $\overline{S_{actual}}$, $\overline{S_{predicted}}$ are actual and ground truth predicted mean values respectively.

## 5. Results and Discussion

### 5.1 Estimation of sustainability indicators

We have reported the performance of different machine learning models for soil erosion factor, soil conditioning index, and organic matter factor in Tables 1, 2 and 3 respectively. We have also plotted the model performance comparing ground truth vs. predicted values in Figures 3, 4, and 5, respectively. Of all the models, the RF regression performed well in predicting the three sustainability indicator factors. Linear regression performed poorly when modeling the three sustainability indicators, as the model failed to capture the nonlinear relationship between the dependent sustainability indicator variables and the independent input variables.

**Table 1:** Performance of different machine learning models to predict soil erosion factors

| | **Performance Metric** | | | | | |
|---|---|---|---|---|---|---|
| | Training Set | | | Test Set | | |
| **Model** | $R^2$ | RMSE | MAE | $R^2$ | RMSE | MAE |
| **Linear Regression** | 0.753 | 0.262 | 0.165 | 0.767 | 0.265 | 0.162 |
| **Ridge Regression** | 0.762 | 0.265 | 0.167 | 0.729 | 0.254 | 0.165 |
| **MLP Regression** | 0.999 | 0.005 | 0.004 | 0.999 | 0.008 | 0.006 |
| **KNN** | 0.886 | 0.182 | 0.092 | 0.838 | 0.198 | 0.099 |
| **SVM** | 0.979 | 0.0455 | 0.0367 | 0.9763 | 0.050 | 0.041 |
| **Decision Tree** | 0.999 | 0.003 | 0.0001 | 0.957 | 0.109 | 0.052 |
| **Gradient Boosting** | 0.969 | 0.092 | 0.056 | 0.965 | 0.098 | 0.057 |
| **Random Forest** | 0.999 | 0.0008 | 0.0001 | 0.999 | 0.003 | 0.0003 |

**Table 2:** Performance of different machine learning models to predict soil conditioning index

| | **Performance Metric** | | | | | |
|---|---|---|---|---|---|---|
| | Training Set | | | Test Set | | |
| **Model** | $R^2$ | RMSE (%) | MAE (%) | $R^2$ | RMSE (%) | MAE(%) |
| **Linear Regression** | 0.875 | 0.071 | 0.051 | 0.876 | 0.0761 | 0.052 |
| **Ridge Regression** | 0.877 | 0.072 | 0.051 | 0.868 | 0.070 | 0.051 |
| **MLP Regression** | 0.992 | 0.018 | 0.023 | 0.976 | 0.031 | 0.0133 |
| **KNN** | 0.908 | 0.062 | 0.042 | 0.871 | 0.073 | 0.051 |
| **SVM** | 0.947 | 0.047 | 0.037 | 0.944 | 0.048 | 0.037 |
| **Decision Tree** | 0.999 | 0.004 | 0.0004 | 0.961 | 0.041 | 0.022 |
| **Gradient Boosting** | 0.981 | 0.029 | 0.020 | 0.975 | 0.029 | 0.020 |
| **Random Forest** | 0.996 | 0.012 | 0.007 | 0.977 | 0.031 | 0.018 |

**Table 3:**
Performance of different machine learning models for predicting Organic Matter Factor

| | **Performance Metric** | | | | | |
|---|---|---|---|---|---|---|
| | Training Set | | | Test Set | | |
| **Model** | $R^2$ | RMSE (%) | MAE (%) | $R^2$ | RMSE (%) | MAE (%) |
| **Linear Regression** | 0.878 | 0.113 | 0.079 | 0.892 | 0.107 | 0.078 |
| **Ridge Regression** | 0.880 | 0.112 | 0.079 | 0.883 | 0.111 | 0.077 |
| **MLP Regression** | 0.986 | 0.038 | 0.026 | 0.973 | 0.055 | 0.037 |
| **KNN** | 0.892 | 0.105 | 0.072 | 0.816 | 0.143 | 0.096 |
| **SVM** | 0.956 | 0.067 | 0.054 | 0.954 | 0.071 | 0.056 |
| **Decision Tree** | 0.999 | 0.007 | 0.0005 | 0.960 | 0.065 | 0.023 |
| **Gradient Boosting** | 0.985 | 0.040 | 0.025 | 0.979 | 0.046 | 0.027 |
| **Random Forest** | 0.996 | 0.020 | 0.009 | 0.968 | 0.058 | 0.026 |

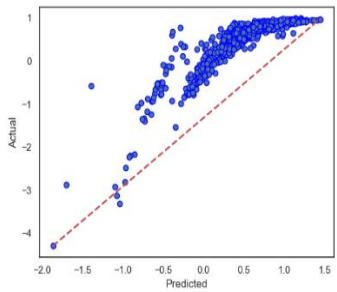

(a) Linear Regression

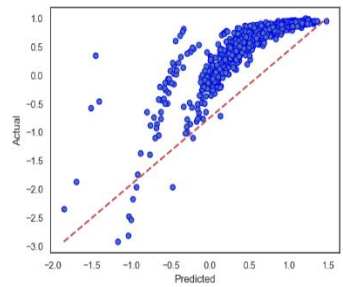

(b) Ridge Regression

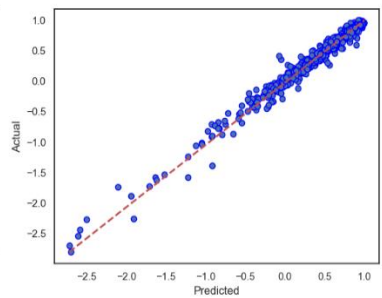

(c) MLP

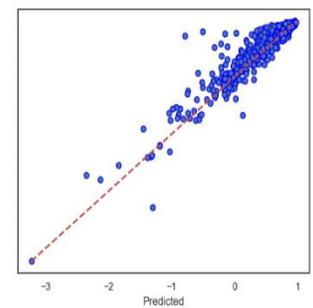

(d) KNN

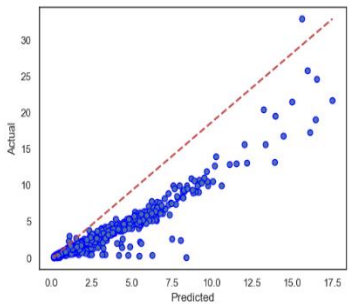

(e) SVM

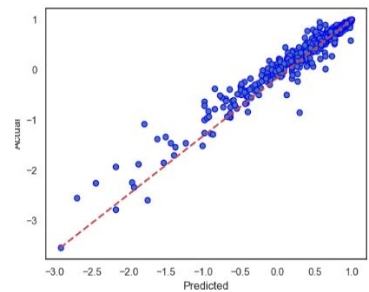

(f) Decision Tree

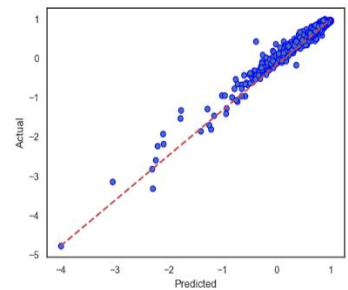

(g) Gradient Boosting

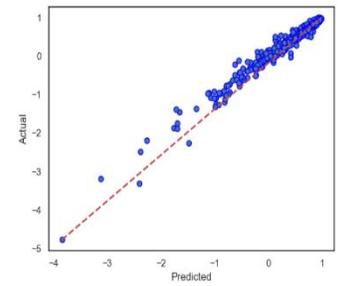

(g) Random Forest

Figure 3: Model Performance to Predict Soil Erosion Factor (SEF)

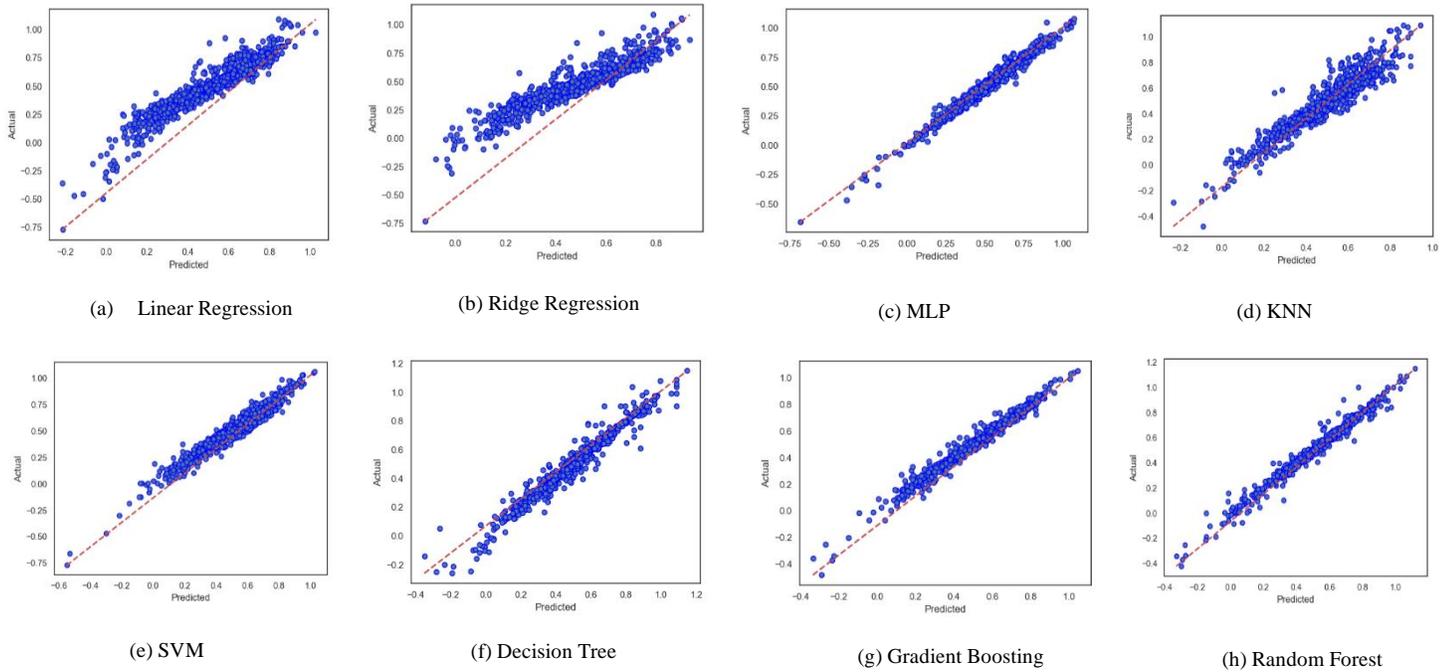

Figure 4: Model performance to Predict Soil Conditioning Index (SCI)

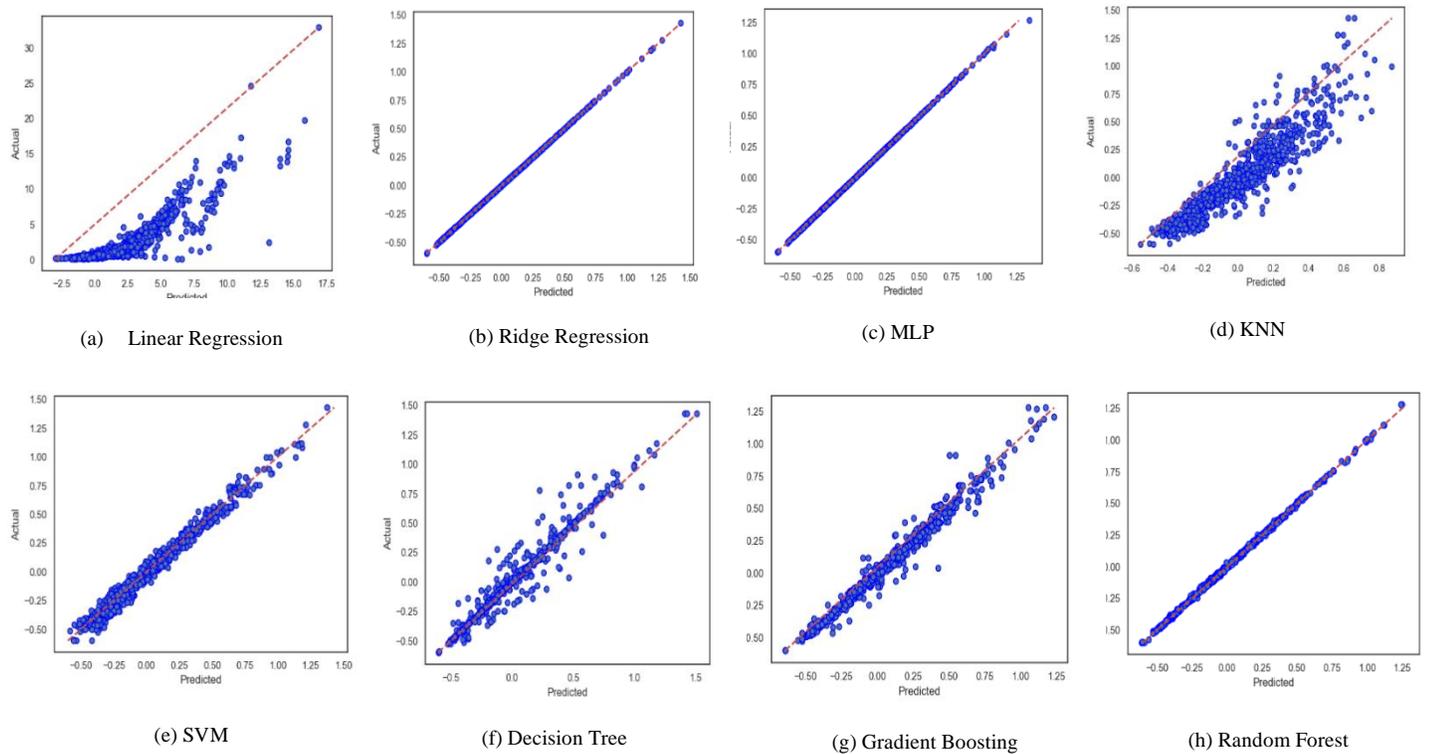

Figure 5: Model Performance to Predict Organic Matter Factor (OMF)

To examine the performance of the voting-based approach in sustainability prediction, we combined strong independent regressors. Based on the reported results in Fig. 3, Fig. 4, and Fig. 5, we combined the Decision Tree, Gradient Boosting, and Random Forest models to compute the average performance. We also experimented with stacking models where a stack of weak learners was bundled together to make a strong predictor model. In the proposed stacked model, the base learner had linear regression, MLP regression, KNN, SVM, and decision tree, whereas the final estimation was performed using the combined model using gradient boost and Random Forest. As seen in Tables 1, 2, and 3, k-NN and linear regression performed poorly to predict sustainability indicators. So, we combined these two models in a stack-wise fashion to make a robust model. Although the voting-based and stacked generalization model did not perform well in this case, better performance was shown for other datasets. Fig. 6 shows the comparative performance of all the models including the assembling models.

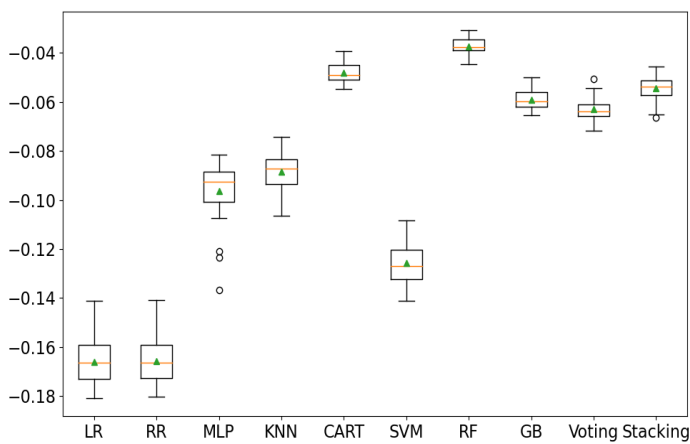

(a) Box plot for stacked model and voting model performance with independent model performance for Soil Erosion Factor (SEF)

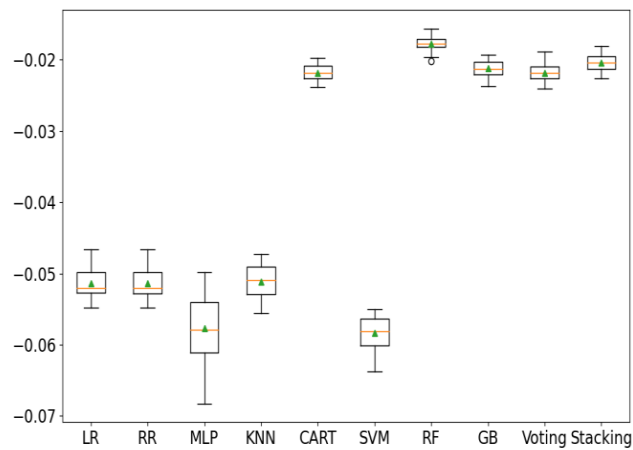

(b) Box plot for stacked model and voting model with independent model performance for Soil Conditioning Index (SCI)

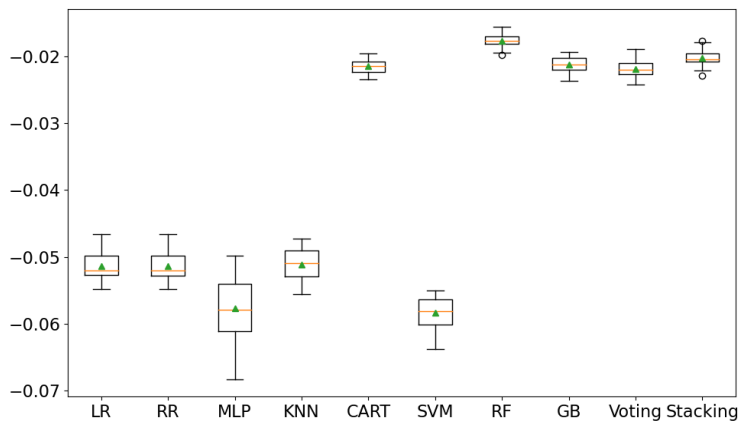

(c) Box plot for the performance of the stacked model with the and the voting model performance of the independent model for Organic Matter Factor (OMF)

Figure 6: Comparative model performance in terms of negative mean absolute error for different sustainability indicator factors.

## 5.2 Estimation of the Residue Removal Rate (RRR)

We have also estimated the performance of the predictive models for residue removal rate (RRR) based on the three sustainability indicator factors, SEF, SCI, and OMF, respectively. Sahoo et al. [11] performed a similar experiment in the same dataset. As shown in Tables 4-6, the best performing model to predict RRR in terms of SEF was Random Forest, which gave an increase of 5% in prediction performance compared to the results reported in [11]. In Table 5, the MLP regression model gave the best results in predicting RRR concerning SCI with high predictability ($R^2$=0.99) compared to the reported results ($R^2$= 0.87) [7]. The Random Forest model better predicted the RRR for OMF ($R^2$=0.99) compared to the results reported in previous studies using ANN [11]. The high predictability of our developed model ensures that the nonlinearity of the response variables is better captured as illustrated in Fig. 7. We can also use the trained model to assess which features are important for RRR prediction. As seen in Fig. 8, we find that soil conditioning index, organic matter factor, clay, and corn yield are the most important features for an accurate estimation of RRR. We have also plotted a comparative model performance in Fig. 9 to predict RRR using different machine learning models.

**Table 4:** Performance of different machine learning models to predict the rate of residue removal with respect to Soil Erosion Factor

| | **Performance Metric** | | | | | |
|---|---|---|---|---|---|---|
| | Training Set | | | Test Set | | |
| **Model** | $R^2$ | RMSE (%) | MAE (%) | $R^2$ | RMSE (%) | MAE (%) |
| **Linear Regression** | 0.375 | 0.222 | 0.177 | 0.348 | 0.232 | 0.186 |
| **Ridge Regression** | 0.877 | 0.072 | 0.0515 | 0.868 | 0.071 | 0.051 |
| **MLP Regression** | 0.943 | 0.068 | 0.050 | 0.884 | 0.096 | 0.068 |
| **KNN** | 0.447 | 0.209 | 0.167 | 0.251 | 0.248 | 0.198 |
| **SVM** | 0.806 | 0.124 | 0.093 | 0.766 | 0.136 | 0.102 |
| **Decision Tree** | 1.0 | 1.984e-17 | 6.16e-18 | 0.727 | 0.147 | 0.081 |
| **Gradient Boosting** | 0.981 | 0.029 | 0.020 | 0.975 | 0.029 | 0.020 |

| | | | | | |
|---|---|---|---|---|---|
| **Random Forest** | 0.997 | 0.016 | 0.006 | 0.977 | 0.042 | 0.017 |

**Table 5:** Performance of different machine learning models for predicting RRR with respect to SCI

| | Performance Metric | | | | | |
|---|---|---|---|---|---|---|
| | Training Set | | | Test Set | | |
| **Model** | $R^2$ | RMSE (%) | MAE (%) | $R^2$ | RMSE (%) | MAE (%) |
| **Linear Regression** | 0.802 | 0.126 | 0.099 | 0.801 | 0.127 | 0.100 |
| **Ridge Regression** | 0.803 | 0.126 | 0.099 | 0.795 | 0.127 | 0.099 |
| **MLP Regression** | 0.989 | 0.029 | 0.022 | 0.976 | 0.045 | 0.032 |
| **KNN** | 0.748 | 0.143 | 0.113 | 0.606 | 0.174 | 0.137 |
| **SVM** | 0.956 | 0.059 | 0.049 | 0.947 | 0.065 | 0.051 |
| **Decision Tree** | 0.999 | 0.002 | 4.13e-05 | 0.882 | 0.096 | 0.039 |
| **Gradient Boosting** | 0.942 | 0.067 | 0.052 | 0.937 | 0.073 | 0.055 |
| **Random Forest** | 0.993 | 0.023 | 0.013 | 0.944 | 0.066 | 0.036 |

**Table 6:** Performance of different machine learning models for predicting residue removal rate with respect to organic matter factor

| | Performance Metric | | | | | |
|---|---|---|---|---|---|---|
| | Training Set | | | Test Set | | |
| **Model** | $R^2$ | RMSE (%) | MAE (%) | $R^2$ | RMSE (%) | MAE (%) |
| **Linear Regression** | 0.838 | 0.114 | 0.088 | 0.846 | 0.109 | 0.086 |
| **Ridge Regression** | 0.842 | 0.112 | 0.086 | 0.829 | 0.117 | 0.088 |
| **MLP Regression** | 0.991 | 0.026 | 0.019 | 0.976 | 0.045 | 0.032 |
| **KNN** | 0.791 | 0.129 | 0.103 | 0.696 | 0.157 | 0.125 |
| **SVM** | 0.959 | 0.057 | 0.048 | 0.951 | 0.061 | 0.049 |
| **Decision Tree** | 1.0 | 8.89e-16 | 3.86e-16 | 0.963 | 0.054 | 0.011 |
| **Gradient Boosting** | 0.970 | 0.049 | 0.033 | 0.959 | 0.058 | 0.038 |
| **Random Forest** | 0.997 | 0.016 | 0.006 | 0.977 | 0.044 | 0.016 |

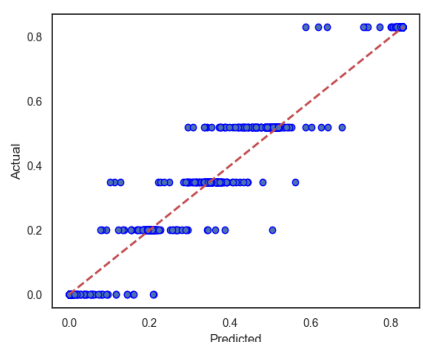 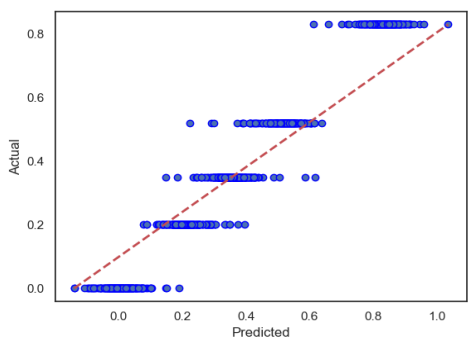 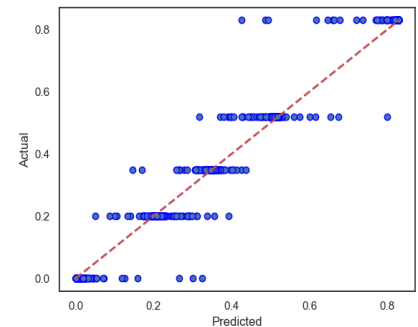

(a) Random Forest    (b) MLP    (c) Random Forest

Figure 7: Best-performing models for predicting residue removal rate with respect to (a) Soil erosion factor, (b) Soil Conditioning Index and (c) Organic Matter Factor

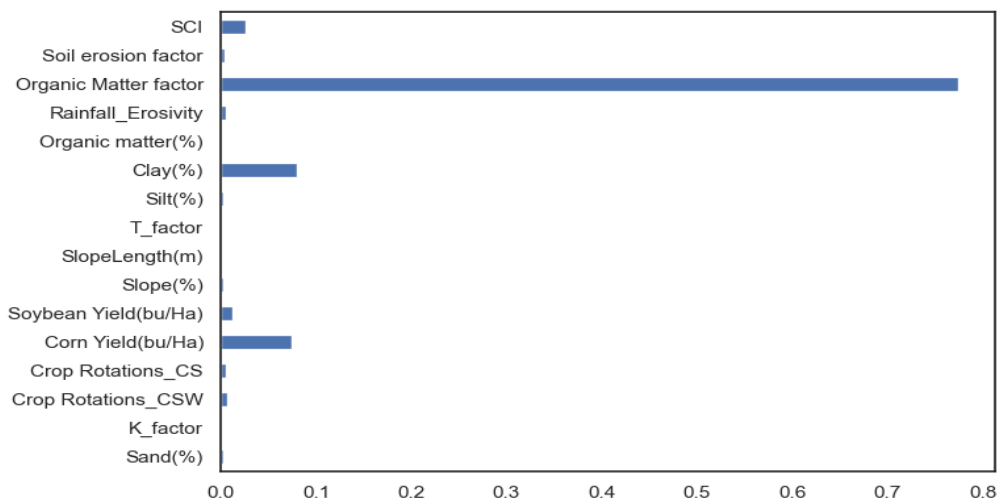

Figure 8: Assessment of the importance using machine learning model to predict the rate of residue removal.

Table 7 delineates the impact of data samples splitting. Random forest performs the best for 70:30 data splitting while for 60:40 data splitting, MLP regression is the best performing model. We have used $R^2$ metric for performance evaluation.

**Table 7:** Impact of splitting factor for predicting residue removal rate with respect to organic matter factor

| Model | 70:30 | | 60:40 | |
|---|---|---|---|---|
| | Training Set | Test Set | Training Set | Test Set |
| **Linear Regression** | 0.839 | 0.842 | 0.839 | 0.839 |
| **Ridge Regression** | 0.840 | 0.837 | 0.842 | 0.835 |
| **MLP Regression** | 0.989 | 0.970 | 0.988 | 0.967 |
| **KNN** | 0.779 | 0.675 | 0.770 | 0.632 |
| **SVM** | 0.959 | 0.944 | 0.957 | 0.948 |
| **Decision Tree** | 1.0 | 0.949 | 1.0 | 0.945 |
| **Gradient Boosting** | 0.969 | 0.962 | 0.972 | 0.964 |
| **Random Forest** | 0.996 | 0.972 | 0.996 | 0.966 |

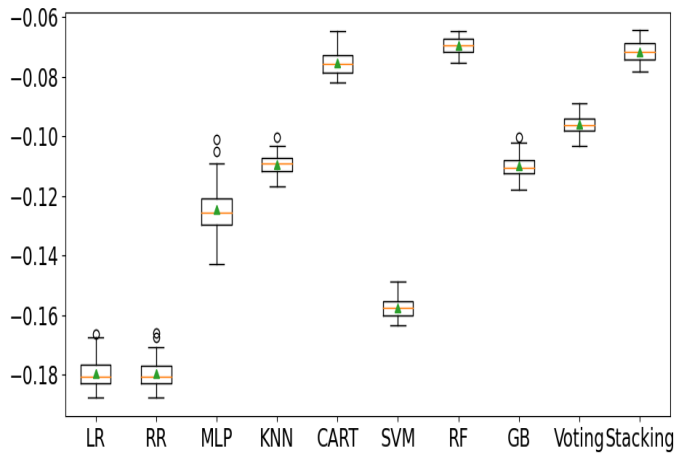

(a) Box plot for stacked model and voting model performance with independent model performance for predicting RRR with respect to Soil Erosion Factor (SEF)

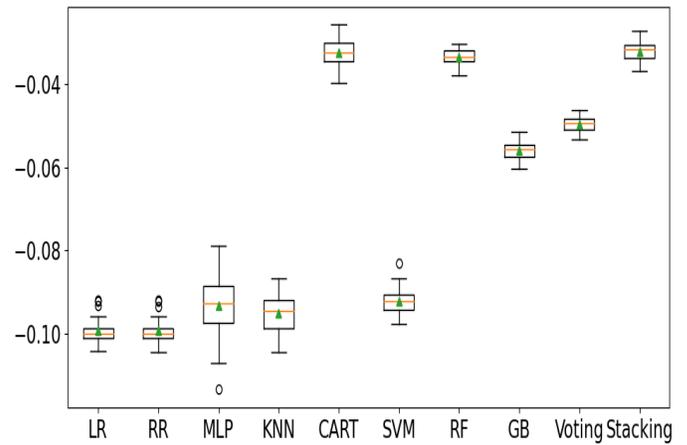

(b) Box plot for stacked model and voting model performance with independent model performance for predicting RRR with respect to Soil Conditioning Index (SCI)

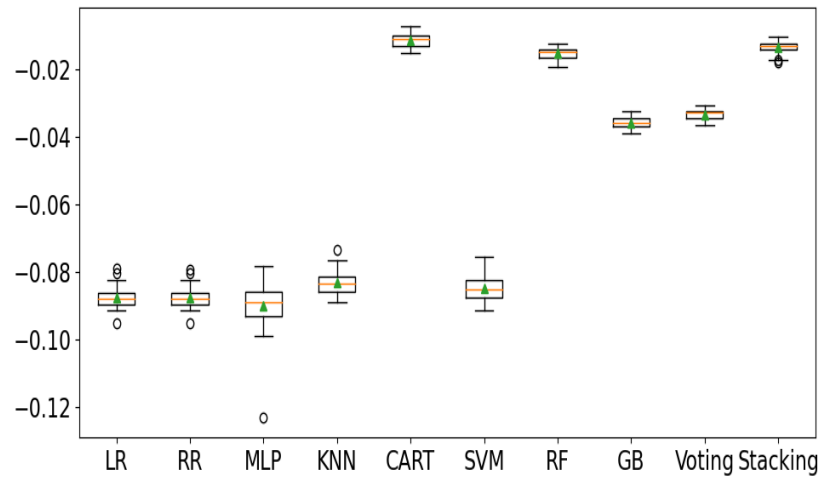

(c) Box plot for stacked model and voting model performance with independent model performance for predicting RRR with respect to Organic Matter Factor (OMF)

Figure 9: Performance of the comparative model in terms of negative mean absolute error in predicting residue removal rate of residue.

# 6 Conclusions

In this work, we investigated the performance of different machine learning approaches for the estimation of three indicators of biomass sustainability using crop residue data. Eight independent machine learning models and two developed ensemble learning based model using stacking and voting approach has been investigated using Ohio State data. The performance of ensemble models has also been investigated for this purpose. Among all the models, Random Forest had the highest goodness of fit, ensuring reliable performance for biomass sustainable indicator prediction. The study has also revealed the importance of different input features for predicting biomass sustainability indicators. The models analyzed can now be integrated with the GIS platform in real time so that the model could be used as a decision support tool in the plant biomass industry.

**Acknowledgement**

This work is supported by Agriculture and Food Research Initiative Competitive Grant no. 2020-68012-31881 from the USDA National Institute of Food and Agriculture. Authors of this paper are employees of the United States Department of Agriculture, U.S. Forest Service. The findings and conclusions in this report are those of the author(s) and should not be construed to represent any official USDA or U.S. Government determination or policy. This research was supported in part by the U.S. Department of Agriculture, Forest Service. Any use of trade, firm, or product names is for descriptive purposes only and does not imply endorsement by the U.S. government.

**Conflicts of Interest**

The authors declare no conflict of interest. The funders of this study had no role in the design of the study; in the collection, analyses, or interpretation of data; in the writing of the manuscript, or in the decision to publish the results.